\documentclass{article}

\makeatletter
\@ifundefined{EX@genericex@mple}{}{\endinput}
\@ifundefined{unitindent}%
  {\newdimen\unitindent\settowidth{\unitindent}{(99)\space a.\space }}%
  {}%
\@ifundefined{ef}{\let\ef\it}{}
\newcommand{\defaultalphex}{%
    \def\examplemakelabel{\rm(\theequation)\hfill}
    \def\alphexmakelabel{\rm\thealphex.\hfill}
    \def\nalphexmakelabel{\rm\thealphex.\hfill}
  }

\defaultalphex
\newcommand{\examplesep}{\abovedisplayskip}
\newcommand{\defaultalphskip}{\def\alphexsep{\jot}\def\alphexlinesep{\jot}}

\defaultalphskip
\newcommand{\alphexskip}{.5\unitindent}
\newcommand{\exampleskip}{\unitindent}
\newcommand{\emptyexskip}{0pt}
\newcommand{\nalphexskip}{\nalphexskipreg}\newskip\nalphexskipreg
\newcounter{alphex}\def\thealphex{\alph{alphex}}
\newenvironment{example}{\begin{EX@genericex@mple}%
  {\examplemakelabel}{\exampleskip}{\emptyexskip}{equation}{}{\exampleskip}}%
 {\end{EX@genericex@mple}\global\@ignoretrue}%
\newenvironment{example*}{\begin{EX@genericex@mple}%
    {\hfill}{\exampleskip}{\emptyexskip}{}{}{\exampleskip}}%
 {\end{EX@genericex@mple}\global\@ignoretrue}
\newenvironment{alphex}{\begin{EX@genericex@mple}%
  {\alphexmakelabel}{\exampleskip}{\alphexskip}{alphex}{new series}{}}%
 {\end{EX@genericex@mple}\global\@ignoretrue}
\newenvironment{calphex}{\begin{EX@genericex@mple}%
  {\alphexmakelabel}{\exampleskip}{\alphexskip}{alphex}{}{}}%
 {\end{EX@genericex@mple}\global\@ignoretrue}
\newenvironment{alphex*}{\begin{EX@genericex@mple}%
  {\hfill}{\exampleskip}{\alphexskip}{}{new series}{}}%
 {\end{EX@genericex@mple}\global\@ignoretrue}
\newenvironment{nalphex}{{\settowidth{\skip0}{\examplemakelabel}%
    \def\@temp{\skip1=}\expandafter\@temp\exampleskip%
    \addtolength{\skip1}{-\skip0}\global\nalphexskipreg\skip1}%
  \begin{EX@genericex@mple}{\nalphexmakelabel}{\exampleskip}{\emptyexskip}%
    {alphex}{new series}{\nalphexskip}}%
 {\end{EX@genericex@mple}\global\@ignoretrue}
\newenvironment{cnalphex}{{\settowidth{\skip0}{\examplemakelabel}%
    \def\@temp{\skip1=}\expandafter\@temp\exampleskip%
    \addtolength{\skip1}{-\skip0}\global\nalphexskipreg\skip1}%
  \begin{EX@genericex@mple}{\nalphexmakelabel}{\exampleskip}{\emptyexskip}%
    {alphex}{}{\nalphexskip}}%
 {\end{EX@genericex@mple}\global\@ignoretrue}
{\catcode`\*=11
\global\let\nalphex*\example*\global\let\endnalphex*\endexample*
\global\let\calphex*\alphex*\global\let\endcalphex*\endalphex*
\global\let\cnalphex*\nalphex*\global\let\endcnalphex*\endnalphex*
\global\let\ncalphex*\cnalphex*\global\let\endncalphex*\endcnalphex*
\global\let\ncalphex\cnalphex\global\let\endncalphex\endcnalphex}
\newif\ifEX@outerex\EX@outerextrue
\def\EX@ifempty#1#2#3{\edef\@temp{#1}\ifx\@temp\@empty#2\else#3\fi}
\newenvironment{EX@genericex@mple}[6]%
 {\ef\def\\{\EX@xstarbr}%
  \begin{list}{{#1}}%
    {\def\@empty{}% defined by LaTeX, but you never know...
     \ifEX@outerex                         % == outer ==
        \topsep\examplesep \itemsep\alphexsep
        \leftmargin#2 \labelwidth\leftmargin
      \else                                % == inner ==
        \topsep\alphexsep \itemsep\alphexlinesep
        \leftmargin#3 \EX@ifempty{#6}{\labelwidth\leftmargin}{\labelwidth#6}
     \fi                                   % == ===== ==
     \partopsep0pt \parsep0pt \labelsep=0pt%
     \EX@ifempty{#4}%
        {\@nmbrlistfalse}%
        {\EX@ifempty{#5}{\@nmbrlisttrue\def\@listctr{#4}}% equals \usecounter
                      {\usecounter{#4}}}%                 \& no counter reset
     \EX@outerexfalse% From now on we are in inner environment
    }%
  \EX@nonumitem}%
 {\end{list}\global\@ignoretrue}
\@ifundefined{@eqpen}{\newcount\@eqpen}{}%again, LaTeX should define \@eqpen.
\newcommand{\EX@xstarbr}{\unskip\@ifstar%
                {\global\@eqpen\@M\EX@ystarbr}%
                {\global\@eqpen\interdisplaylinepenalty \EX@ystarbr}}
\newcommand{\EX@ystarbr}{\@ifnextchar [{\EX@zstarbr}{\EX@zstarbr[\z@]}}
\def\EX@zstarbr[#1]{\penalty\@eqpen\vskip#1\EX@nonumitem}
\def\EX@nonumitem{\expandafter\EX@uxnonumitem\EX@ident}% gobble all spaces
\def\EX@uxnonumitem{\bgroup\def\nonumber{\nonumber}%
\def\@tempb{\nonumber}\futurelet\@tempa\EX@nextn@n}%
\def\EX@nextn@n{\ifx\@tempa\@tempb\gdef\@tempa{\item[]}%
              \else\gdef\@tempa{\item\relax}\fi\egroup\@tempa}
\long\def\EX@ident#1{#1}% as \@iden in \LaTeX but \long, this doesn't help,
\newcommand{\addprime}{\global\advance \c@alphex by 26\relax}
\newcommand{\subtrprime}
           {\ifnum\c@alphex>26\global\advance \c@alphex by -26\relax\fi}
\newcommand{\resetalphex}
           {\@ifnextchar[{\EX@reset@lphex}{\global\c@alphex=0\relax}}
\newcommand\oldstylealph{\let\alph\@@alph\let\Alph\@@Alph}
\newcommand\newstylealph{\let\alph\EX@alph@i\let\Alph\EX@Alph@i}
\newcommand{\lastref}{\@ifnextchar[{\EX@lastref}{\arabic{equation}}}
\newcommand{\lastalph}{\@ifnextchar[{\EX@lastalph}{\alph{alphex}}}
\newcount\c@EX@aa@cnt\c@EX@aa@cnt=0
\def\EX@lastref[#1]{\c@EX@aa@cnt=#1\c@EX@aa@cnt=-\c@EX@aa@cnt
        \advance \c@EX@aa@cnt by \c@equation \arabic{EX@aa@cnt}}
\def\EX@lastalph[#1]{\c@EX@aa@cnt=#1\c@EX@aa@cnt=-\c@EX@aa@cnt
        \advance \c@EX@aa@cnt by \c@alphex \alph{EX@aa@cnt}}
\newcommand{\EX@alph@i}[1]{\EX@alph{\@nameuse{c@#1}}}
\newcommand{\EX@Alph@i}[1]{\EX@Alph{\@nameuse{c@#1}}}
\let\@@alph\alph\let\@@Alph\Alph
\newstylealph
\def\EX@reset@lphex[#1]{\global\c@alphex=#1\relax}
\newcommand{\EX@alph}[1]%
  {\ifcase#1\or a\or b\or c\or d\else\EX@ialph{#1}\fi}
\newcommand{\EX@ialph}[1]%
  {\ifcase #1\or \or \or \or \or e\or f\or g\or h\or i\or j\or
   k\or l\or m\or n\or o\or p\or q\or r\or s\or t\or u\or v\or
   w\or x\or y\or z\else\EX@acc{#1}{}\EX@alph\fi
  }
\newcommand{\EX@Alph}[1]%
  {\ifcase#1\or A\or B\or C\or D\else\EX@iAlph{#1}\fi}
\newcommand{\EX@iAlph}[1]%
  {\ifcase #1\or \or \or \or \or E\or F\or G\or H\or I\or J\or
   K\or L\or M\or N\or O\or P\or Q\or R\or S\or T\or U\or V\or
   W\or X\or Y\or Z\else\EX@acc{#1}{}\EX@Alph\fi
  }
\newcommand{\EX@acc}[3]%
 {\ifnum#1<27%
   {\setbox0\hbox{#3{#1}}\hbox to \wd0{$\hbox{\box0}#2$\hss}}%
  \else
   \c@EX@aa@cnt=#1\advance \c@EX@aa@cnt by -26\relax
   \EX@acc\c@EX@aa@cnt{#2'}{#3}%
  \fi
 }
\makeatother

\newcommand{\bc}{\begin{center}}
\newcommand {\ec}{ \end{center}}
\newcommand {\be} {\begin{enumerate}}
\newcommand {\ee} {\end{enumerate}}
\newcommand {\bi} {\begin{itemize}}
\newcommand {\ei} {\end{itemize}}
\newcommand {\ba} {\begin{array}}
\newcommand {\et} {\end{tabular}}
\newcommand {\bt} {\begin{tabular}}
\newcommand {\ea} {\end{array}}
\newcommand {\bex} {\begin{example}}
\newcommand {\eex} {\end{example}}
\newcommand {\bsex} {\begin{alphex}}
\newcommand {\esex} {\end{alphex}}

\newcommand {\txt}[1]{\mbox{{ \it{#1}}}}
\def\wff{\txt{wff}}

\def\Covars{{\cal C\kern-.2ex V}}

\def\evnup{\@ifnextchar[{\@evnup}{\@evnup[0pt]}}
\def\@evnup[#1]#2{\setbox1=\hbox{#2}%
\dimen1=\ht1 \advance\dimen1 by -.5\baselineskip%
\advance\dimen1 by -#1%
\leavevmode\lower\dimen1\box1}

\begin{document}

%%--------------------------------------

\title{Corrections and Higher--Order Unification}

\author{Claire Gardent, Michael Kohlhase, Noor van Leusen}

\date{}

\maketitle

\thispagestyle{empty}

%%--------------------------------------

\begin{abstract}
\noindent
We propose an analysis of corrections which models
some of the requirements corrections place on context. We then show
  that this analysis naturally extends to the interaction of
  corrections with pronominal anaphora on the one hand, and
  (in)definiteness on the other. The analysis builds on previous
  unification--based approaches to NL semantics and relies on
  Higher--Order Unification with Equivalences, a form of unification
  which takes into account not only syntactic $\beta\eta$-identity but also
  denotational equivalence.
\vspace{2ex}

\noindent
Wir schlagen eine Analyse vor, die einige der
Anforderungen von Korrekturen an den Kontext modelliert und
sich nat\"urlich auf die Interaktion von Korrekturen mit
Pronominalanaphern und Undefiniertheit erweitern l\"a{\ss}t.  Die
Analyse basiert auf bekannten unifikationsbasierten Ans\"atzen f\"ur
die Semantik nat\"urlicher Sprache und benutzt eine Erweiterung der
Unifikation h\"oherer Stufe. Diese ber\"ucksichtigt nicht nur
strukturelle $\beta\eta$-Gleichheit, sondern auch logische
\"Aqui\-va\-lenz.

\end{abstract}

%%--------------------------------------

\setlength{\parindent}{0em}
\setlength{\parskip}{1ex}

%%--------------------------------------
%%--------------------------------------

\section{Introduction}
\label{s1}
Corrections are utterances such as (\ref{ex:e1}b) where a
discourse participant corrects the utterance of some other
discourse participant\footnote{Here and in what follows, we use
  capital letters to indicate prosodic prominence.}.

\bex
\label{ex:e1}
\bsex
A: Jon likes Mary. \\
B: No, PETER likes Mary.
\esex
\eex

Although there is much literature on corrections (e.g.
\cite{ScJeSa:tpfscitooric77,norrick:otooceic91,ringlebruce:sfnlp82}),
a thorough investigation of their linguistics is still outstanding. In
this paper, we build up on \cite{Leusen:tioc94} and examine some of
the requirements corrections place on context or in other words, the
relationship between correction (the correcting utterance) and
correctee (the utterance being corrected). For instance, it is clear
that the pair of utterances in (\ref{ex:e2}) does not form a
well--formed dialog.

\bex
\label{ex:e2}
\bsex
A: Jon likes Mary. \\
$\star$ B: No, PETER likes Sarah.
\esex
\eex

On the other hand, it is also clear that a simple equality requirement
between the semantic representation of the deaccented part of the
correction and that of its parallel counterpart in the source is not
appropriate either:

\bex
\label{ex:e3}
\bsex
A: Jon likes [the woman with the red hat]$_1$
\\
B: No, PETER likes  Sarah$_1$
\esex
\eex

Here the correction contains an NP {\it Sarah} whose semantic
representation is not identical with that of its source parallel
element {\it the woman with the red hat}. In other words, a
requirement such as \cite{Sag:dalf76}'s alphabetical variant
constraint would fail\footnote{Sag proposes an analysis of VP ellipsis
  which requires that the semantic representation of a VP ellipsis be
  an alphabetical variant of the semantic representation of its
  antecedent. The basic assumption is that semantic representations
  are $\lambda$--terms. Two terms are alphabetical variants of each
  other iff they are identical up to renaming of bound variables.}. At
this stage one could be tempted to conclude that the equality
requirement is a semantic one: the deaccented part of the correction
must be semantically equivalent with its parallel correlate in the
source utterance. However, this is also incorrect. Thus in
(\ref{ex:e4}), the property denoted by the VP in (\ref{ex:e4}b) need
not be the same as the property denoted by its parallel counterpart in
(\ref{ex:e4}a): whereas the VP in (\ref{ex:e4}a) denotes the property
of loving Jon's wife, the VP in (\ref{ex:e4}b) may denote the property
of loving Peter's wife\footnote{This is of course similar to the
  sloppy/strict ambiguity characteristic of VP ellipsis. Indeed, as we
  shall later see, our treatment is very similar to
  \cite{DaShPe:eahou91}'s treatment of VPE.}.

\bex
\label{ex:e4}
\bsex
A: Jon$_1$ loves his$_1$ wife. \\
B: No, PETER loves his wife.
\esex
\eex

In short, it is clear that some identity requirement is needed to
appropriately characterise the relation between correctum and
correction (cf. example \ref{ex:e2}). On the other hand, it is less
clear what this identity requirement should be (cf. examples
\ref{ex:e3},\ref{ex:e4}). In this paper, we contend that the correct
notion of identity is given by Higher--Order Unification with
equivalences, a form of Unification which takes into account not only
syntactic identity, but also denotational equivalence.  We show that
the HOUE--based analysis of corrections we propose, not only captures
some of the contextual requirements of corrections, but also makes
appropriate predictions about the interaction of corrections with both
pronominal anaphora and (in)definiteness.

\section{HOU with Equivalences}
\label{s2}

Now we will briefly review higher-order unification and its properties, for
details we refer the reader to~\cite{Snyder91}. Higher-order unification
solves the problem of finding substitutions $\sigma$ that for a given
equation $A=B$ make both sides equal in the theory of $\beta\eta$-equality
($\sigma(A)=_{\beta\eta}\sigma(B)$). Huet's well-known
algorithm~\cite{Huet75} solves the problem by recursively decomposing
formulae and binding Function variables to most general formulae of a
given type and given head.

However, even though HOU considers $\beta\eta$-equality of formulae, it
does not take into account the semantics of the logical connectives and
quantifiers contained in the logical representation of natural language
utterances. For this we need a unification algorithm for
$\beta\eta$-equality augmented by logical equivalence. Obviously, such an
algorithm has to generalize theorem proving methods for higher-order logic,
since the task of unifying an equation $(A\vee\neg A) = T$, where $T$ is a
sentence, is equivalent to proving the validity of the theorem
$T$\footnote{The formula $(A\vee\neg A)$ must be true in all models, so $T$
  can only be equivalent to it, if it is a theorem.}. An algorithm that
solves this problem is described in~\cite{Kohlhase:hot95}. It is a
generalization of the first-order Tableau method~\cite{Fitting:fotp} for
automated theorem proving, which refutes a negated theorem by analyzing the
connectives in an and/or tree and finding instantiations that close each
branch of the tree by finding elementary contradictions on it.

Instead of a formal recapitulation of the tableau method, we discuss the
example of the logical theorem $(p(a)\vee p(b)\Rightarrow \exists x.p(x))$.
The negation of this is equivalent\footnote{In addition to the de Morgan
laws we use the identity $\exists x.A=\neg\forall x.\neg A$.} to the
formula at the root of the following tableau.
\[\begin{array}{c}
  p(a)\vee p(b) \wedge \forall x.\neg p(x) \\
  p(a)\vee p(b) \\
  \forall x.\neg p(x) \\
  \left.
    \begin{array}{c}
   p(a) \\{}
   \neg p(y) \\{}
    *[y=a]
    \end{array}\right|
  \begin{array}{c}
    p(b) \\{}
   \neg p(z) \\{}
    *[z=b]
  \end{array}
\end{array}\]
Here we see that conjuncts are simply added to the branch, whereas
disjuncts
 are analyzed in separate branches of the tree. The scopes of
universal quantifications (with new variables) can be inserted at the end of
branches, the same is possible with the scopes of existential quantifications
(with the bound variables replaced by Skolem\footnote{Skolem terms serve as
  witnesses for the objects whose existence is claimed by the existential
  formula $A$. Since this object may depend on the values of free variables
  $x_1,\ldots,x_n$ occurring in $A$, they have the form $f(x_1,\ldots,x_n)$
  where $f$ is a new function.} terms). Finally, both branches of the tableau
are closed, i.e. the last formula can be instantiated (by the substitution in
brackets) so that it contradicts a formula in the branch above.

These instantiations are computed by unification, and in the case of
higher-order logic by HOU. The distinguishing feature of the HOUE
algorithm~\cite{Kohlhase:hot95} is that intermediate equations $(A =
B)$ of type $t$ (generated either by unifying two formulae on the
branch to make them contradictory or by processing other unification
problems) can be transformed into negated equivalences (which can then
be treated by the theorem proving component). Actually, tableau
development for the negated equivalence $\neg(A\Leftrightarrow B)$
contains trivial branches, so we use the following (optimized) rule,
which splits an equation of type $t$ into two tableau branches
\[
\begin{array}{c}
   A=B \\
\begin{array}{c|c}
   A      & B \\
   \neg B & \neg A
\end{array}\end{array}\]
This way, HOU and tableau theorem proving recursively call each other in
HOUE, until a refutation is found (all branches of the tableau are closed).

\section{The basic analysis}
\label{s3}

Typically, a correction partially or completely repeats a previous
utterance and one of its characteristic properties is that the
repeated material is deaccented, that is, it is characterised by
an important reduction in pitch, amplitude and duration (cf.
\cite{Bartels:sot95}).  Our proposal is to analyse corrections as
involving a deaccented anaphor which consists of the repeated
material. Furthermore, we require that the semantic representation of
a deaccented anaphor unify with the semantic representation of its
antecedent.

More precisely, let $SSem$ and $TSem$ be the semantic
representations of the source (i.e. antecedent) and target (i.e.
anaphoric) clause respectively, and $TP^1 \dots TP^n, SP^1 \dots SP^n$
be the target and source parallel elements\footnote{As in
  \cite{DaShPe:eahou91}, we take the identification of parallel elements as
  given.}, then the interpretation of an SOE must respect the
following equations: \\ 

$
\ba{l}
An(SP^1,\dots,SP^n) = SSem \\
An(TP^1,\dots,TP^n) = TSem
\\
\\
\ea
$

Intuitively, these two equations require that target and source clause
share a common semantics: $An$, the semantics of the deaccented
anaphor. We illustrate the workings of the analysis by a simple
example. Given the dialog in (\ref{ex:e1}), the equations to be solved
are:
\\

$
\ba{l}
An(j) = like(j,m) \\
An(p) = like(p,m)
\\
\\
\ea
$

Given these equations, HOU yields a unique solution $An = \lambda x.
like(x,m)$. In contrast, the equations required for the analysis of
example (\ref{ex:e2}) are: \\ 

$
\ba{l}
An(j) = like(j,m) \\
An(p) = like(p,s)
\\ \\
\ea
$

Since there is no substitution of values for free variables which
simultaneously makes $An(j)$ $\alpha\beta\eta$--identical with
$like(j,m)$ and $An(p)$ $\alpha\beta\eta$--identical with $like(p,s)$,
unification fails thereby indicating the ill--formedness of (\ref{ex:e2}).

\section{Corrections and pronominal anaphora}
\label{s4}

The resolution of pronouns occurring in the destressed part of a
correction appears to be subject to very strong parallelism
constraints. For instance in (\ref{ex:e10}b), the pronoun {\it her}
can only be understood as referring to its source parallel element {\it Sarah} 
-- else it must be stressed. 

\bex
\label{ex:e10}
\bsex
Jon loves Sarah$_1$ .
\\
No, PETER loves her.
\esex
\eex

Intuitively, there is a simple explanation for this: if the destressed
part of a correction is a repeat of its parallel element in the source
utterance, then pronouns occurring in it must necessarily resolve to
their parallel counterpart in the source expression. As we shall see,
the picture is somewhat more complex however. In some cases, a
destressed pronoun in the correction may be ambiguous. In other cases,
it functions as a paycheck pronoun. Finally, extraneous factors such
as scope constraints and world knowledge interact with the semantics
of corrections in determining the resolution of destressed pronouns.
In what follows, we show how HOUE allows us to correctly predict this
array of empirical facts.

 \subsection{Pronouns}

 Let us start with example (\ref{ex:e10}) above.  Given
 the analysis of corrections described in section \ref{s3}, the
 equations to be resolved are\footnote{Unresolved pronouns are
   represented by free variables i.e. variables whose value is determined
   by unification. Alternatively, pronouns could be resolved first and
   unification would then function as a filter on admissible
   resolutions.}:
\\

$
\begin{array}{ll}
An(j) & = \txt{love}(j, s) \\
An(p) & = \txt{love}(p, x) \\
&
\end{array}
$

By unification, the only possible values for $An$ and $x$ are
$\lambda y\txt{love}(y, s)$ and $s$ respectively. That is, the
destressed pronoun is resolved by unification to its parallel element
in the source utterance, {\it Sarah}. As required.

In some cases however, a destressed pronoun in the correction is
ambiguous. For instance in (\ref{ex:e6}b), the pronoun {\it his} may
resolve either to {\it Jon} or to {\it Peter}.

\bex
\label{ex:e6}
\bsex
Jon$_1$ loves his$_1$ wife.
\\
No, PETER $_2$ loves his$_{1,2}$ wife.
\esex
\eex

Interestingly, such cases are similar to the sloppy/strict
ambiguity\footnote{The terminology {\it sloppy/strict} originated with
  \cite{Ross:covis67}. Intuitively, a pronoun has a strict
  interpretation if it denotes as its antecedent. By contrast, a
  pronoun which denotes differently from its antecedent is said to
  have a sloppy interpretation.} characteristic of VP ellipsis and as
\cite{DaShPe:eahou91} have shown, HOU straightforwardly captures such
cases because of its ability to yield several solutions. In the case
of (\ref{ex:e6}), the analysis proceeds as follows. First, the
following equations must be resolved: \\ 

$
\begin{array}{ll}
An(j) & = \txt{love}(j, \txt{wof }(j)) \\
An(p) & = \txt{love}(p, \txt{wof }(x)) \\
&
\end{array}
$

Resolution of the first equation yields two values for
$An$\footnote{Unification yields a third value for $An$, namely
  $\lambda y\txt{love}(j, \txt{wof}(y))$. This solution however is ruled
  out by the second equation. More generally, we assume a restriction
  similar to \cite{DaShPe:eahou91}'s {\bf Primary Occurrence
    Restriction} (POR): the occurrences directly associated with the
  contrastive elements are primary occurrences and any solution
  containing a primary occurrence is discarded as linguistically
  invalid. For instance, in $An(j) = \txt{love}(j, \txt{wof }(j))$,
  the first occurrence of $j$ is a primary occurrence so that the
  solution $An = \lambda y$ {\it love(j, wof(y))} is ruled out. For a
  proposal of how the POR can be formally modelled, see
  \cite{GaKo:hocuanls96}.}, $\lambda y \txt{love}(y, \txt{wof}(j))$
and $\lambda y \txt{love}(y, \txt{wof}(y))$. By applying $An$ to $p$,
we then get two possible values for $An(p)$: $\txt{love}(p,
\txt{wof}(j))$ and $\txt{love}(p, \txt{wof}(p))$. As a side
effect, the pronoun {\it his} represented by $x$ is resolved either to
{\it Jon} or to {\it Peter}. In short, for such cases, the multiple
solutions delivered by HOU match the ambiguity of natural
language.
 
\subsection{Paycheck pronouns}

Destressed pronouns whose source parallel element is a pronominal
possessive NP are particularly interesting. At first sight, they seem
to behave just like any other destressed pronouns occurring in a
correction, that is, they seem to resolve unambiguously to their parallel
source element. For instance, in (\ref{ex:e14}b), the most likely
resolution of {\it her} is {\it Jon's wife}.

\bex
\label{ex:e14}
\bsex
Jon$_1$ likes his$_1$ wife. 
\\
No, PETER likes her (= his$_1$ wife)
\esex
\eex

However, a closer investigation of the data suggests that this reading is a kind 
of default reading which is preferred out of a pair of two grammatically 
possible interpretations. To see this, consider examples (\ref{ex:e15}) and 
(\ref{ex:e16}).
\bex
\label{ex:e15}
\bsex
Jon$_1$ broke his$_1$ arm yesterday.
\\
No, PETER$_2$ broke it (= his$_{1,2}$ arm) yesterday.
\esex
\eex

\bex
\label{ex:e16}
\bsex
Jon$_1$ had his$_1$ nose remodelled in Paris.
\\
No, PETER$_2$ had it (= his$_{2}$ nose) remodelled in Paris.
\esex
\eex

Although these examples are structurally identical with
(\ref{ex:e14}), they differ in the interpretation of the destressed
pronoun occurring in the correction. Whereas (\ref{ex:e14}) only
allows for a strict interpretation of this pronoun,
(\ref{ex:e15}) permits both a strict and a sloppy interpretation whilst
(\ref{ex:e16}) only admits of a sloppy reading.

Our contention is that a destressed pronoun in the correction whose
source parallel element is a possessive definite, is systematically
ambiguous between a strict and a sloppy interpretation.  However
extraneous factors may have the effect that only one reading is
available. For instance, in (\ref{ex:e16}) the strict reading is ruled
out by our world knowledge that one can only have one's own nose
remodelled. As for (\ref{ex:e14}), the absence of sloppy reading can
be explained if we assume that the interpretation of a destressed
anaphor follows a default strategy geared toward maximal semantic
identity between the destressed anaphor and its antecedent. Under this
assumption, the strict reading is the most natural since it
establishes a strict denotational identity between the antecedent VP
{\it likes Jon's wife} and the destressed anaphor {\it likes her}.

The behaviour of these pronouns is simply explained once they are
viewed as {\it paycheck pronouns} as illustrated by Karttunen's famous
example (cf. \cite{karttunen:pav69}):

\bex
\label{ex:e30}
The man who gave his paycheck to his wife was wiser than the man who
gave it to his mistress
\eex

Paycheck pronouns differ from other pronouns in that they can neither
be seen as coreferential constants nor as bound variables -- instead
they pick up the definite description introduced by their
antecedent and reanchor its possessive pronoun in its immediate
context. For instance in (\ref{ex:e30}) above, the paycheck pronoun
{\it it} picks up the description {\it his paycheck} and reanchors its
possessive pronoun {\it his} to the second occurrence of {\it the
  man}. 

There are various ways in which paycheck pronouns can be accounted for
but essentially, the idea is that their denotation is fixed by a
definite description containing either an unresolved pronoun or an
unresolved property. As \cite{cooper:tiop79} convincingly argues, the
second solution is methodologically more satisfactory. We will
therefore assume that paycheck pronouns are definite NPs whose
representation includes a free variable of type $(e \rightarrow t)$
i.e. a property. More specifically, we assume that a paycheck pronoun
is assigned the following representation:

\[
\lambda Q. \exists x[P(x) \wedge \forall
y[P(y) \leftrightarrow y = x] \wedge Q(x)]
\]

\noindent
where $P\in\wff_{(e \rightarrow t)}$. Given this, the analysis of
(\ref{ex:e14}) runs as follows. The equations to be resolved to check
the well-formedness of the destressed anaphor {\it likes her}
are:\footnote{In what follows, we abbreviate $\lambda Q. \exists
  x[P(x) \wedge \forall y[P(y) \leftrightarrow y = x] \wedge Q(x)]$ to
  $\lambda Q. \exists x[P(x) \wedge unique(x) \wedge Q(x)]$.}.

$
\begin{array}{ll}
An(j) & = \exists x [\txt{wof}(x, j) \wedge unique(x)\wedge love(j,x)] \\
An(p) & = \exists x [P(x) \wedge unique(x) \wedge love(p,x)]
\end{array}
$

Resolution of the first equation yields the two values $\lambda y.
\exists x [\txt{wof}(x, j) \wedge unique(x) \wedge \txt{love}(y,x)]$
and $\lambda y.  \exists x [\txt{wof}(x, y) \wedge unique(x)\wedge
\txt{love}(y,x)]$ for $An$ and thus, the values $\exists x
[\txt{wof}(x, j) \wedge unique(x)\wedge\txt{love}(p,x)]$, and $\exists x
[\txt{wof}(x, p)\wedge unique(x) \wedge \txt{love}(p,x)]$ for $An(p)$.
The first result yields the strict reading (Peter loves Jon's wife)
whereas the second yields the sloppy reading (Peter loves Peter's
wife).

\section{Corrections and definiteness}
\label{s5}

So far, we have only considered cases where the semantic
representation of the destressed anaphor could syntactically unify
with that of its antecedent. That is, in each case it was possible to
find a substitution of values for free variables which made the two
semantic representations $\alpha\beta\eta$--identical.  In this
section, we turn to more semantic cases, cases in which the relation
between destressed anaphor and source parallel element is one of
denotational -- rather than syntactical -- identity. Definites are a
primary example of such a phenomenon: since one and the same
individual can be referred to by several, distinct definite
descriptions, it often happens that the definite description used in
the destressed part of a correction is not structurally identical with
the description used in its source parallel element. This is
illustrated in example (\ref{ex:e20}) where the source utterance
contains the definite {\it the woman with the red hat}. As illustrated
by (\ref{ex:e20}a--d), the parallel element in the correction can be
{\it his wife, her, the neighbour's daughter} or {\it Sarah}. In each
case, the description does not syntactically unify with the source
description {\it the woman with the red hat}. Note however that the
correction is only well--formed when the parallel descriptions are
interpreted as referring to one and the same individual (cf. the
ill--formedness of (\ref{ex:e20}e--g)). That is, when they are
semantically equivalent.

 \bex
\label{ex:e20}
Jon$_2$ likes [the woman with the red hat]$_1$
\bsex
No, PETER$_3$ likes his wife (= NP$_1$) \\
No, PETER likes her$_1$.\\
No, PETER likes [the neighbour's daughter]$_1$. \\
No, PETER likes Sarah$_1$. \\
$\star$ No, PETER likes her$_4$.\\
$\star$ No, PETER likes Mary$_4$. \\
$\star$ No, PETER likes him. 
\esex
\eex

How does HOUE account for such examples? To show this, we now sketch
the main steps of the unification process for example (\ref{ex:e20}d)
with equations:
\[\begin{array}{ll}
An(p) & =   like(p,s) \\
An(j) & = \exists x (w(x) \wedge wrh(x) \wedge unique(x) \wedge like(j,x)) 
\end{array}\]
These are solved in a context, where Sarah is the only woman with a
red hat. The HOUE method is given access to the hypotheses
$unique(s)$, $w(s)$ and $wrh(s)$ by adding them to the initial
tableau.  In a first step, we solve the first equation to
$An=\lambda z.like(z,s)$ and obtain the following tableau:
\[\begin{array}{c} 
 unique(s)\\
 w(s)\\ 
 wrh(s)\\
 An(p)  =   like(p,s) \\
  \vdots\\
   like(j,s)=\exists x (w(x)\wedge wrh(x)\wedge unique(x) \wedge like(j,x))
  \end{array}\]
The HOUE rule discussed in section~\ref{s2} now splits the initial
equation into two branches.  The first one has the form 
\[\begin{array}[t]{c}
        like(j,s) \\
        \neg\exists x
        (w(x)\wedge\ldots \wedge like(j,x))\\
        \begin{array}{c|c|c|c}
          \neg w(z) & \neg wrh(z) &\neg unique(z) &\neg like(j,z) \\{}
          *[z=s]    &  *[z=s]     & *[z=s]        &   *[z=s]    
        \end{array}
    \end{array}\]
and contains the formulae $like(j,s)$ and
$(\neg\exists x (w(x)\wedge wrh(x)\wedge unique(x) \wedge
like(j,x)))$. The latter is universally quantified\footnote{We use
  that $\neg\exists x.A$ is equivalent to $\forall x.\neg A$ here.}
and can therefore be developed into four branches $\neg w(z)$, $\neg
wrh(z)$, $\neg unique(z)$, and $\neg like(j,z)$. The first three branches
can be closed using the hypotheses on Sarah and the last one with the
first formula, all by binding the new variable $z$ to $s$.  The
second branch has the form 
    \[\begin{array}[t]{c}
       \neg like(j,s)\\
       \exists x (\ldots \wedge unique(x)\wedge like(j,x))\\
       unique(c)\\
       like(j,c))\\
       \vdots \\
       c=s \\
       like(j,s)\\{}
       *[]
    \end{array}\]
and consists of the formulae $\neg like(j,s)$ and $\exists
x (w(x)\wedge wrh(x)\wedge unique(x) \wedge like(j,x))$, which is
developed into the single branch containing the conjuncts $w(c)$,
$wrh(c)$, $unique(c)$, and $like(j,c))$, where $c$ is a Skolem
constant for $x$. Here an expansion of the definition of uniqueness
\[unique(x)\Leftrightarrow\forall z(w(z)\wedge wrh(z)\leftrightarrow x=z)\]
closes the branch (if Sarah and $c$ are unique, then $s=c$).

By now, it should be clear that our treatment will also encounter no
particular problem in dealing with examples such as (\ref{ex:e23}) and
(\ref{ex:e24}) below.  The first example relies on the world-knowledge that
marrying is a symmetric relation (both partners have to say ``yes I
do''), whereas the second relies on the fact that getting wounded is
synonymous to being hurt by someone/thing. Once these equivalences are
taken into account, the HOUE analysis of corrections will correctly
predict that these examples are well--formed.

\bex
\label{ex:e23}
\bsex
A: Jon married Sarah \\
B: No, Sarah married PETER
\esex
\eex

\bex
\label{ex:e24}
\bsex
A: Sarah hurt Paul. \\
B: No, PETER was wounded.
\esex
\eex

We have seen that a deaccented anaphor must either have a semantic
representation which syntactically unifies with that of its antecedent, or be
semantically equivalent to this antecedent. To show that this is a necessary
condition, we need to provide some ill--formed examples in which neither
condition holds. Such examples are given when the correction contains a
destressed pronoun whose source parallel element is either an indefinite
(\ref{ex:e21}) or a quantifier (\ref{ex:e22}).

\bex
\label{ex:e21}
\bsex
Jon eats an$_1$ apple.
\\
$\ast$ No, PETER eats it$_1$. 
\esex
\eex

\bex
\label{ex:e22}
\bsex
Jon kissed most$_1$ women at the party yesterday.
\\
$\ast$ No, PETER kissed them$_1$. 
\esex
\eex

In both cases, the semantic representation of the pronoun in the
correction fails to syntactically unify with the semantic
representation of its antecedent. Neither can it be proved that {\it
  it} and {\it them} are semantically equivalent to {\it an apple} and
{\it most women at the party } respectively. Therefore, unification
fails correctly ruling out (\ref{ex:e21}) and (\ref{ex:e22}). The
logical reason for this e.g. in (\ref{ex:e21}), is that while the
second equation $An(p)=eat(p,y)$ can be solved to $An=\lambda
x.eat(x,y)$ yielding the negated $\neg(eat(j,y)\Leftrightarrow\exists
x (ap(x)\wedge eat(j,x))$, this cannot be refuted
\footnote{Example (\ref{ex:e21}) is in fact ambiguous between a specific
  reading of the indefinite {\it an apple} and a non-specific one. In
  the first case, the indefinite denote uniquely so that {\it it} in
  (\ref{ex:e21}b) refers to this unique apple. Since it is
  denotationally equivalent with its antecedent, HOUE will succeed. In
  the second case, there is no unique apple salient in the context,
  hence {\it it} and {\it an apple} cannot be denotationally
  equivalent. Therefore HOUE fails. The above discussion focuses on
  this second possibility.}.

\section{Conclusion}
\label{s6}

In a sense, it would be much more natural to express the proposed
analysis in a dynamic setting (cf. \cite{Kamp:atotasr81}). The data
discussed in section \ref{s5} clearly shows that definite, indefinites
and quantifiers behave differently wrt. corrections. The intuition is
that whereas, a definite can bind a pronoun in the correction (cf.
example \ref{ex:e20}), indefinites and quantifiers cannot (cf.
examples \ref{ex:e21},\ref{ex:e22}). These are of course precisely the
sort of facts dynamic semantics was designed to deal with: if we
assume that the correctee--correction pair is semantically represented
by a disjunction $(\Phi \vee \Psi)$, then a definite in the correctee
will be able to bind an anaphor in the correction (because definites
have global scope) whereas indefinites and quantifiers won't
(because traditionally disjunction is static and the discourse
referents introduced by one disjunct are not accessible to the other
disjunct).  In this paper, we've shown that such facts could be
modelled by means of HOUE on static semantic representations; it would
be interesting to see how the analysis would transpose to a more
dynamic setting. This however must await the development of
Higher--Order Unification for a dynamic lambda--calculus.

Another question worth investigating is whether the interleaving of
anaphora resolution and quantification proposed in
\cite{DaShPe:eahou91} could account for the data considered here. The
approach has the advantage that it does not resort to equivalences,
thus permitting better computational properties. However, unless
definites are treated in a special way, it is unlikely that the
approach will be able to capture examples such as (\ref{ex:e20}) where
denotational equivalence, rather than strict unification, is
required. 

Finally, an interesting issue concerns the relationship between HOUE
and accommodation. A simple way to model accomodation would be to posit
that, as theorem proving hits a dead-end, accomodation can be used to
close off a branch: the accomodated fact is the fact needed to derive
a contradiction and close off this tableau branch. Naturally, this
idea is too simplistic in that some model must be defined which
constrains accomodation. This we leave as an open research issue.

\end{document}